\documentclass[11pt]{article}

\usepackage[preprint]{acl}

\usepackage{times}
\usepackage{latexsym}
\usepackage{graphicx}
\usepackage{subcaption} 
\usepackage[T1]{fontenc}

\usepackage[utf8]{inputenc}

\usepackage{microtype}

\usepackage{inconsolata}

\usepackage{graphicx}

\usepackage{multirow}
\usepackage[table]{xcolor}
\usepackage{array} 
\usepackage[most]{tcolorbox}
\newcolumntype{g}{>{\columncolor{gray!20}}c}
\newcolumntype{b}{>{\columncolor{blue!20}}c}
\usepackage{colortbl}
\colorlet{gr}{gray!15}
\colorlet{bl}{blue!15}
\colorlet{gr2}{gray!30}
\colorlet{bl2}{blue!30}
\usepackage{listings}
\newcommand{\red}{\color{red}}
\newcommand{\blue}{\color{blue}}

\title{{Do LLMs Recognize Your Latent Preferences? A Benchmark for Latent Information Discovery in Personalized Interaction}}

\author{Ioannis Tsaknakis$^{1}$, \; {\bf Bingqing Song}$^{1}$, \; Shuyu Gan$^{1}$, \; {\bf Dongyeop Kang}$^{1}$ \\
        {\bf Alfredo Garcia$^{2}$, \; Gaowen Liu$^{3}$, \; Charles Fleming$^{3}$, \; Mingyi Hong$^{1}$} \\
        $^{1}$ University of Minnesota, \; $^{2}$ Texas A\&M University, \; $^{3}$CISCO Research}

\begin{document}
\maketitle

\begin{abstract}

Large Language Models (LLMs) excel at producing broadly relevant text, but this generality becomes a limitation when user-specific preferences are required, such as recommending restaurants or planning travel. 
In these scenarios, users rarely articulate every preference explicitly; instead, much of what they care about remains \emph{latent}, waiting to be inferred. 
This raises a fundamental question: \textit{Can LLMs uncover and reason about such latent information through conversation?}

We address this problem by introducing a unified benchmark for evaluating \textit{latent} information discovery - the ability of LLMs to reveal and utilize hidden user attributes through multi-turn interaction. The benchmark spans three progressively realistic settings: the classic {20 Questions} game, {Personalized Question Answering}, and {Personalized Text Summarization}. 
All tasks share a tri-agent framework (User–Assistant–Judge) enabling turn-level evaluation of elicitation and adaptation.
Our results reveal that while LLMs can indeed surface latent information through dialogue, their success varies dramatically with context: from 32\% to 98\%, depending on task complexity, topic, and number of hidden attributes. 
This benchmark provides the first systematic framework for studying \textit{latent information discovery} in personalized interaction, highlighting that effective preference inference remains an open frontier for building truly adaptive AI systems.
\end{abstract}

\section{Introduction}\label{sec:intro}
Large Language Models (LLMs) have achieved remarkable success across domains, such as healthcare \cite{wang-etal-2025-survey, busch2025current}, education \cite{chu2025llm, xiao-etal-2023-evaluating},  translation \cite{wang-etal-2024-benchmarking, zhu-etal-2024-multilingual}, and code generation \cite{jiang2024survey, zan-etal-2023-large}.
Modern LLMs can now produce responses that are accurate, coherent, and contextually appropriate. 
Yet these strengths do no guarantee \textit{user satisfaction}: many LLM responses remain overly \emph{generic}, designed to appeal to a broad audience rather than reflect an individual's intent or constraints.

As LLMs increasingly mediate personal and professional decision-making (e.g., recommending restaurants, summarizing news, or offering tailored advice), this lack of \textit{personalization} becomes a critical limitation. 
In practice, user-specific information is often \textit{latent} -unstated, contextual, or even subconscious-, and therefore unavailable to the model at the outset.

\noindent \textbf{A simple motivating example.}  
Consider the classic game of \emph{20 Questions}. One player thinks of an object, and the other must infer it by asking a series of yes-or-no questions. 
Success requires \textit{strategic questioning} to reveal hidden information. 
Note replace the hidden object with a user's unstated preference, like a dietary restriction or stylistic goal.
Can an LLM play this game, asking the right questions to infer what the user values and tailoring its final answer accordingly? 
This analogy illustrates the core challenge of this work: \textit{effective personalization requires latent information discovery.}

\begin{figure}[t!]
    \centering
    \includegraphics[width=1\linewidth]{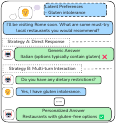}
    \caption{Illustration of a question answering task. Strategy A: The LLM directly provides a response. This response is generic and violates the user's preferences. Strategy B: The LLM tries to elicit the user preferences through a multi-turn conversation, requesting specific information at each turn. At the end of the interaction, the LLM produces a personalized response.
    }
    \label{fig:intro_ex}
    \vspace{-0.5cm}
\end{figure}

\paragraph{How can LLMs personalize?}  
We can imagine three possible strategies: {\bf (1)} User explicitly provides all preferences in the prompt; {\bf (2)} LLM requests them before responding; {\bf  (3)} LLM actively \emph{elicits} them through {\it multi-turn conversation} — asking targeted questions, and interpreting answers, and adapting its final response.

While the first two strategies are technically viable, they are rarely realistic: users seldom recall or articulate every relevant preference upfront. The third approach, \textit{interactive elicitation}, is therefore the most natural and potentially powerful path toward genuine personalization.
Fig.~\ref{fig:intro_ex} illustrates this contrast between static and interactive strategies.

\paragraph{Our benchmark.}
Building on this intuition, we design an unified benchmark for evaluating LLMs' ability to discover and use latent user information through multi-turn dialogue.
The benchmark follows a consistent tri-agent framework, comprising a User (with hidden preferences), an Assistant (the model under evaluation), and a Judge (which evaluates alignment), and spans three increasingly realistic tasks:
\textbf{(1)} \emph{20 Questions}: a controlled reasoning setting isolating pure latent information discovery; 
\textbf{(2)} \emph{Personalized Question Answering}: goal-oriented dialogue requiring semantic reasoning about user constraints; and
\textbf{(3)} \emph{Personalized Text Summarization}: document-level synthesis guided by user-specific summarization preferences.
This design enables systematic, turn-level evaluation of questioning strategies, reasoning efficiency, and personalization accuracy. We focus on the passive-user setting, where the user only responds when prompted, mirroring real-world (and in some sense worst case) interactions where users rarely volunteer all relevant context. 

Benchmarking open- and closed-source models on our dataset reveals performance varying from 32–98\%, depending on task complexity, topic, and number of hidden attributes.
Our results also show that LLMs can partially infer latent information but struggle when cues are sparse or multi-dimensional, exposing critical gaps in adaptive reasoning.
Overall, we can say that latent preference identification is neither inherently easy nor inherently difficult; instead, its level of difficulty depends on the specific setting and other factors such as the number of latent preferences.
Together, these results establish the first systematic foundation for studying \textit{latent information discovery}, a core yet underexplored skill required for building truly adaptive, user-centered LLMs.

Finally, it is worth noting that our benchmark can be interpreted through the lens of preference tree models in decision theory \cite{tversky1979preference}, where each user’s latent preference structure defines a hierarchical set of aspects. Through dialogue, the LLM effectively performs an elimination over this latent tree, by asking targeted questions to prune inconsistent branches until the user’s intended preference leaf is revealed.

\begin{figure*}
    \centering
    \includegraphics[width=0.7\linewidth]{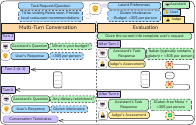}
    \caption{The benchmark’s general setting, including the example of Fig. \ref{fig:intro_ex}. The boxes with a solid border depict the generic elements of the setting, while the boxes with a dashed border describe the example. Both the multi-turn conversation between the Assistant and the User, and the evaluation phase (where the Judge is utilized) after each turn are depicted.}
    \label{fig:setting}
\end{figure*}

\section{Related Works}
\noindent\textbf{Multi-Turn Interaction Benchmarks.}
Several studies have evaluated LLM performance in multi-turn conversations, focusing on general dialogue capabilities rather than personalization. 
For instance, \citet{kwan-etal-2024-mt} assess models’ abilities in \emph{recollection} (retaining information from earlier turns) and \emph{refinement} (revising previous instructions), while \citet{bai-etal-2024-mt} examine \emph{perceptivity}, \emph{adaptability}, and \emph{interactivity} across multi-turn dialogue scenarios. 
Similarly, \citet{zhang-etal-2024-probing} employ the classic \emph{20 Questions} game as a surrogate task to analyze LLMs’ reasoning and planning skills over multiple turns.  
While these studies shed light on the generic dynamics of conversational reasoning, they do not address how models can \emph{actively elicit and integrate hidden user information}, which is the focus of our work. 
In contrast, our benchmark frames multi-turn dialogue as a means for \textbf{latent information discovery}, explicitly testing whether LLMs can uncover and leverage unspoken user attributes.

\noindent\textbf{LLM Personalization Benchmarks.}
A growing body of work investigates how LLMs can adapt to individual users through personalization tasks. 
However, existing benchmarks largely evaluate how models \emph{follow} or \emph{apply} known preferences, not how they \emph{uncover} them from the users. 
For example, \citet{zhaollms} assess preference following and memorization abilities in long-context conversations, while \citet{afzoon2024persobench} evaluate both open- and closed-source LLMs on generating responses consistent with a given persona and conversational context. 
The LaMP \cite{salemi-etal-2024-lamp} and LongLaMP \cite{kumar2024longlamp} benchmarks extend this idea by evaluating model outputs across eleven personalization tasks, such as personalized product rating and email completion. 
More recently, \citet{zhao-etal-2025-personalens} also study personalization through multi-turn interactions, focusing on explicit persona conditioning and interactive adaptation. 

Our work differs from all of the above in both {goal} and {setting}. 
We focus on {\it preference unveiling} rather than {\it preference following}: the model must \emph{discover} hidden user information rather than be told what it is. 
Furthermore, we evaluate this capability under a {\it passive-user assumption}, where the user only responds to model queries, forcing the LLM to decide what to ask and when. 
In addition, our benchmark explores a diverse set of tasks—including the {20 Questions} game, {Personalized Question Answering}, and {Personalized Text Summarization}, so to better connect abstract reasoning with realistic personalization scenarios.

\section{Benchmark Setting and Methodology}

In this section, we present the overall design of our benchmark and the methodology used to evaluate LLMs in latent information discovery. 
We first describe the general setup shared across all tasks, and then detail the task-specific implementations.

\subsection{General Setting and Methodology}

\begin{figure*}
    \centering
    \includegraphics[width=0.9\linewidth]{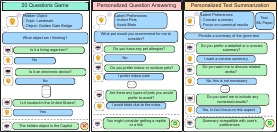}
    \caption{The two personalization tasks and the 20 Questions Game task of our benchmark. The LLM tries to elicit the user preferences (or the hidden object) through a multi-turn conversation, requesting specific information at each turn. After each turn, the Assistant produces a personalized response to the User's task (here we only illustrate the response after the final turn).}
    \label{fig:tasks}
\end{figure*}

The overall benchmark structure is illustrated in Fig.~\ref{fig:setting}, and the three evaluated tasks are shown in Fig.~\ref{fig:tasks}. 
Each task is designed to assess how well an LLM can uncover hidden user information through multi-turn interaction and leverage that information to produce personalized responses.

\noindent\textbf{Tasks.} 
The benchmark centers on tasks that require the discovery of hidden or implicit user information. 
A successful Assistant must go beyond producing a generally correct answer—it must uncover and incorporate the User’s latent preferences through interaction. 
We include three tasks that together capture this spectrum of behavior.

\noindent(1) {\it Personalized Question Answering} and (2) {\it Personalized Text Summarization} are the two core personalization tasks. 
In both, the Assistant must identify which pieces of information about the User are relevant to the current task (e.g., dietary restrictions, stylistic preferences) and then use them to generate a tailored output. 
These tasks mimic realistic user–LLM interactions, where success depends on how effectively the model elicits, interprets, and applies latent user cues. In both task we employ a diverse set of topics to capture different scenarios. Specifically, we have five topics for task (1) and four topics for task (2). 

\noindent (3) {\it The 20 Questions Game} serves as a foundational and diagnostic component of the benchmark. 
Although not a personalization task in the conventional sense, it serves as a foundational and diagnostic component of the benchmark. 
It isolates the \emph{pure reasoning process} of latent information discovery. 
Here, the hidden object plays the same conceptual role as a latent preference: it is information unknown to the Assistant that must be uncovered through strategic questioning. 
This task provides a clean, controlled environment for examining the reasoning mechanisms that underpin successful personalization. 
By abstracting away domain context such as text or user profiles, it focuses solely on the Assistant’s ability to plan, hypothesize, and update beliefs across turns. 
Performance in this setting therefore offers a complementary diagnostic perspective; it illustrates how well a model can reason about and uncover hidden information before those capabilities are tested in more complex, domain-specific personalization tasks.
 
\begin{figure*}[h]
\centering
    \centering
    \includegraphics[width=\linewidth,trim=0 0 0cm 0cm,clip]{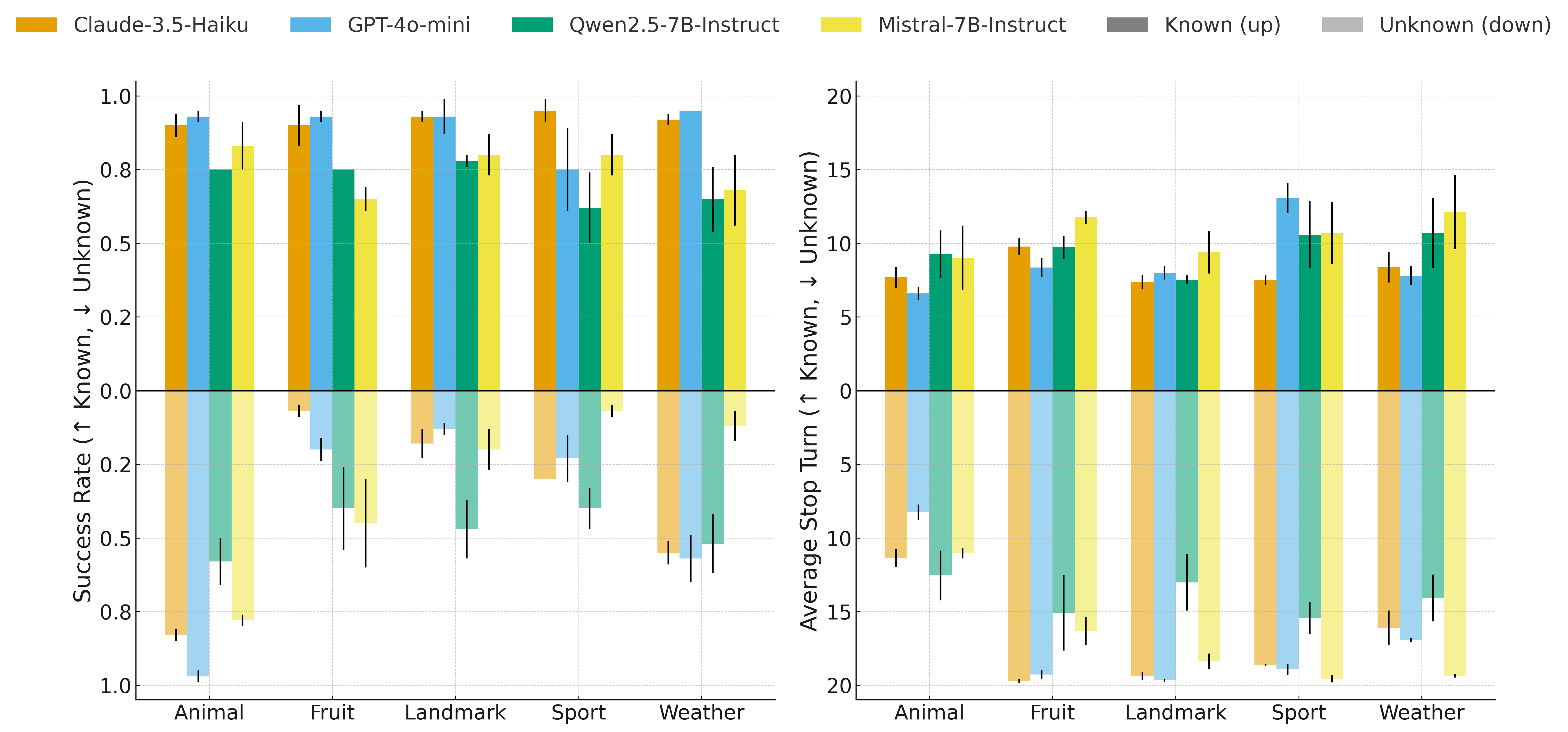}
\caption{Results of the ``20 Questions Game'' experiments across different models, topics, and conditions; Topic Known ↑ (the upper part of each figure) vs Topic Unknown ↓ (the lower part of each figure).
}
\label{fig:20q_combined}
\end{figure*} 
\noindent\textbf{Agents.} 
Each task involves three interacting agents: the \textbf{Assistant}, the \textbf{User}, and the \textbf{Judge}. 
The \textbf{User} simulates a person with fixed latent preferences and provides concise, truthful responses to the Assistant’s questions. 
The \textbf{Assistant} acts as the model under evaluation and is the active participant in the dialogue. 
Its goal is to uncover the User’s hidden information and complete the task in a way that aligns with those preferences. 
After each conversational round, the Assistant generates a task-level output (e.g., an answer or a summary). 
Finally, the \textbf{Judge} operates externally, assessing whether the Assistant’s output satisfies all of the User’s latent preferences. \\

\noindent\textbf{Multi-Turn Interaction.} 
The dialogue unfolds as a sequence of Assistant–User exchanges. 
At each turn, the Assistant poses a targeted question and the User replies according to their latent preferences. 
The Assistant may refine its subsequent questions using all prior context. 
The maximum number of turns is fixed and known to the Assistant in advance, ensuring comparability across models. \\

\noindent\textbf{Evaluation and Early Stopping.} 
After each turn (i.e., one Assistant–User exchange), the Assistant produces a complete, personalized response to the User’s original task using all accumulated information. 
The Judge then evaluates whether this output satisfies every latent preference. 
If alignment is achieved, the conversation terminates early (\emph{success}); otherwise, it continues until the maximum number of turns is reached (\emph{failure}). 
This early-stopping mechanism allows us to measure both the quality and efficiency of latent-information discovery. \\

\noindent\textbf{Evaluation Metrics.} 
We use two primary metrics:
\begin{enumerate}
\vspace{-0.2cm}
    \item \textbf{Success Rate} — the proportion of instances in which the Assistant’s final output satisfies all latent preferences;
    \vspace{-0.2cm}
    \item \textbf{Average Stop Turn} — the average number of turns required to reach success. This metric is computed over all instances by treating the maximum possible value as the stop turn in cases of failure (e.g., 20 for the 20 Questions).
    \vspace{-0.2cm}
\end{enumerate}

Additionally, as part of our evaluation we conduct an error type analysis where we examine the different ways the Assistant models fails. For further details and the results of this analysis, refer to Sec. \ref{sec:error_analysis}. Finally, we conduct an evaluation of our Judge model to examine the extent to which the Judge LLM adheres to its provided instructions. The purpose of this evaluation is to demonstrate that the specific Judge model employed is reliable, thereby ensuring the accuracy of our results. The evaluation of the Judge is provided in Sec. \ref{sec:judge_eval} in the Appendix. \\

\noindent\textbf{Models.} 
All agents are implemented with large language models. 
To ensure reliability, the \textbf{User} and \textbf{Judge} are instantiated using strong, stable models that perform their roles consistently across runs. 
Specifically, we employ \textbf{GPT-4o}~\cite{hurst2024gpt} for both. 
For the \textbf{Assistant}, we evaluate four representative models spanning both closed- and open-source families: 
\textbf{GPT-4o-mini}~\cite{hurst2024gpt}, \textbf{Claude-3.5-Haiku}, \textbf{Mistral-7B-Instruct}~\cite{jiang2023clip}, and \textbf{Qwen2.5-7B-Instruct}~\cite{yang2025qwen3}. 
This diverse model set enables a comparative analysis of how architecture, training strategy, and scale influence latent-information discovery in interactive settings.

\begin{figure*}[t]
\centering
\includegraphics[width=\linewidth]{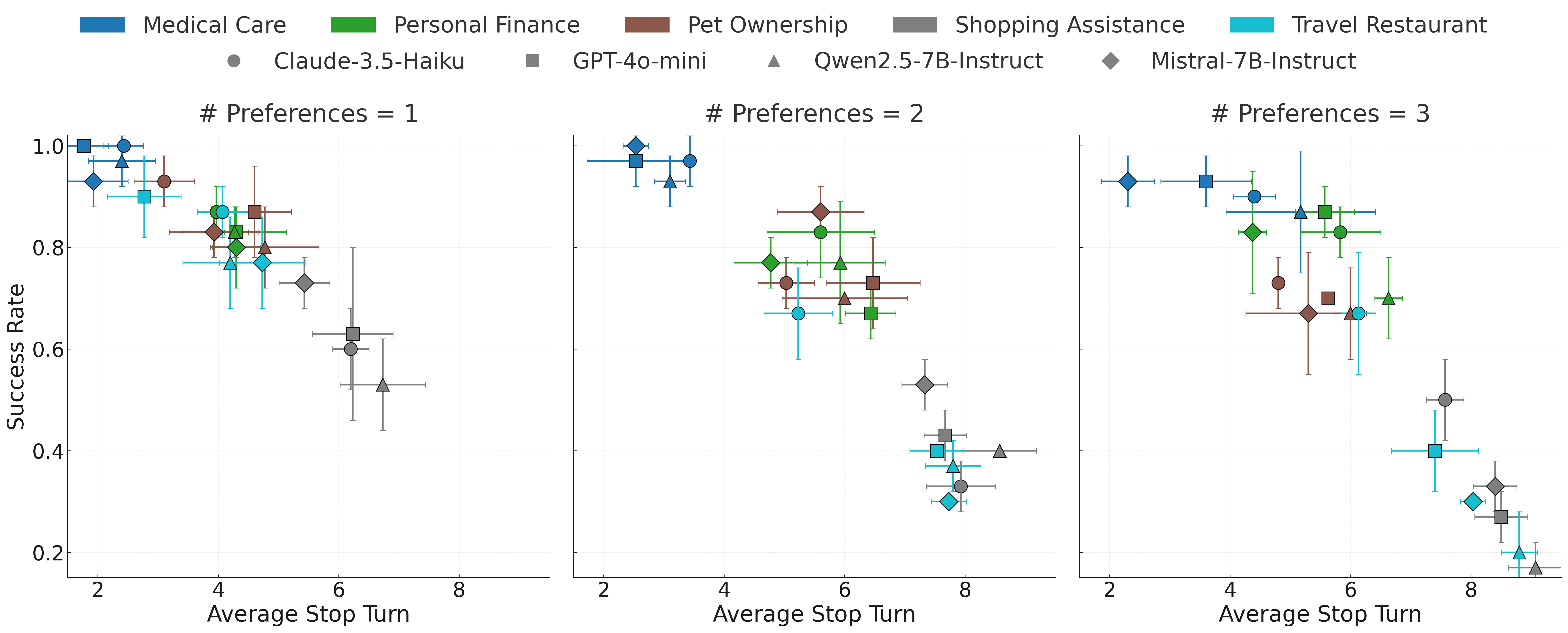}
\caption{Results of the Personalized Question Answering experiments across different models, topics, and preference settings. 
The figure shows Success Rate (y-axis) vs.\ Average Stop Turn (x-axis) for \#Preferences $\in\{1,2,3\}$ (three panels). 
Color encodes \emph{topic/domain} and marker shape encodes \emph{model}; error bars reflect reported uncertainties on both axes.
(i) Domain effects are prominent: clusters by topic are clearly separated; 
(ii) As the number of preferences increases, points generally drift toward the lower-right (lower success, higher turns), indicating that deeper personalization settings tend to be harder. 
}
\label{fig:pqa_scatter_domains_models}
\end{figure*}

\subsection{20 Questions Game}

The first component of our benchmark is the {20 Questions Game}, a structured variant of the classic guessing game in which one participant selects an object and the other attempts to identify it by asking a sequence of yes–or–no questions.  Each \textbf{User} is associated with a single hidden object drawn from one of five topics: \textit{animal}, \textit{fruit}, \textit{landmark}, \textit{sport}, or \textit{weather}. 
The topic may either be disclosed to, or withheld from the \textbf{Assistant}, allowing us to evaluate how prior knowledge affects questioning strategies. 
During the interaction, the Assistant poses one question per turn, and the User responds truthfully with ``yes,'' ``no,'' or ``not applicable.'' 
The dialogue lasts for at most $20$ turns. 

After every turn, the Assistant must, outside of the dialogue, propose its current best guess of the hidden object using all gathered information. 
The \textbf{Judge} then verifies whether this guess is correct. 
If correct, the conversation terminates early (\emph{success}); otherwise, it continues until the $20$-turn limit is reached (\emph{failure}). 

\subsection{Personalized Question Answering}

In Personalized Question Answering, each User is associated with a list of one to three latent preferences and a question. The preferences and the question correspond to a specific topic from the following list: medical care, personal finance, pet ownership, shopping assistance, travel restaurant. 

The Assistant is given the question, and its goal is to provide a personalized answer, i.e., an answer that aligns with all the preferences in the User's list. To achieve this, the Assistant asks the User targeted questions aimed at eliciting all relevant (to the User's question) preferences, specifically a single question per conversation turn to which the User answers truthfully. 

After each turn, we request from the Assistant to provide a response to the User's question (outside of the conversation), using the information gathered so far.  The Judge assesses whether the Assistant's response correctly aligns with all of the User's latent preferences. The maximum number of conversation turns is set to $10$.

\begin{figure*}[ht]
\centering
{\small
    \begin{minipage}[t]{0.34\textwidth}
        \begin{tcolorbox}[colback=black!5, colframe=gray!50, boxrule=0pt, arc=0mm]
        \vspace{-0.5cm}
        Preference: I avoid spicy foods. \vspace{-0.2cm}
        \end{tcolorbox}
    \end{minipage}
    \begin{minipage}[t]{0.65\textwidth}
        \begin{tcolorbox}[colback=green!5, colframe=gray!50, boxrule=0pt, arc=0mm]
        \vspace{-0.5cm}
        Q: Can you recommend some restaurants to try during my visit to Mumbai?\vspace{-0.2cm}
        \end{tcolorbox}
    \end{minipage}
    \\
    \begin{minipage}[t]{1\textwidth}
        \begin{tcolorbox}[colback=blue!5, colframe=gray!50, boxrule=0pt, arc=2mm]
        \vspace{-0.2cm}
        \textbf{GPT-4o-mini}\\
        1.A: {\blue What type of cuisine do you enjoy the most?} \\
        1.U: {\red I enjoy a variety of cuisines, but I especially like Indian, Italian, and Chinese food.}\\
        2.A: {\blue Are you looking for fine dining experiences, casual eateries, or a mix of both during your visit?}\\
        2.U: {\red I'm looking for a mix of both fine dining experiences and casual eateries during my visit.}\\
        3.A: {\blue Do you have any dietary restrictions or preferences that I should consider when recommending restaurants?}\\
        3.U: {\red I avoid eating spicy foods entirely.}\vspace{-0.2cm}
        \end{tcolorbox}
    \end{minipage}
    \\
    \begin{minipage}[t]{1\textwidth}
        \begin{tcolorbox}[colback=gray!5, colframe=gray!50, boxrule=0pt, arc=2mm]
        \vspace{-0.2cm}
        \textbf{Qwen-2.5-7B-Instruct}\\
        1.A: {\blue What kind of cuisine are you particularly interested in trying during your visit to Mumbai?} \\
        1.U: {\red I'm open to trying a variety of cuisines, but I have a particular interest in exploring authentic Indian dishes.}\\
        2.A: {\blue Are there any specific regions or states within India that you're curious to explore through their cuisine?}\\
        2.U: {\red I'm curious to explore the cuisine of Maharashtra. I'd love to try some authentic local dishes from this region.}\\
        3.A: {\blue Are there any specific types of dishes from Maharashtra that you're excited to try, such as vegetarian or seafood?}\\
        3.U: {\red I'm excited to try a variety of dishes from Maharashtra, including both vegetarian and seafood options.}\\
        4.A: {\blue Are there any specific festivals or occasions in Mumbai that you're planning to experience, which might influence the type of food you'd like to try?}\\
        4.U: {\red I'm not visiting during any specific festivals or occasions, so my focus will mainly be on exploring the local cuisine.}\\
        5.A: {\blue Are there any dietary restrictions or preferences you have, such as vegetarian, vegan, or gluten-free options?}\\
        5.U: {\red I avoid eating spicy foods entirely, so I'd appreciate recommendations for restaurants that offer milder options.} \vspace{-0.2cm}
        \end{tcolorbox} 
    \end{minipage}
}
\vspace{-0.7cm}
\caption{An instance of the Personalized Question Answering task for the ``Travel Restaurant topic'', along with the conversations of two models. We note that GPT-4o begins by asking more general questions, allowing it to identify the latent preference (e.g., avoid spicy food) earlier than Qwen-2.5-7B-Instruct. In contrast, the latter model immediately specializes its questions after the first turn (e.g., asking about specific regions) and becomes ``trapped'' in a narrow conversation, thereby postponing the identification of more general preferences.}
\label{fig:qa_example}
\end{figure*}

\subsection{Personalized Text Summarization}

 In this task, each user is associated with a list of three preferences and a text. The preferences and the texts correspond to a specific topic from the following list: news articles (``cnn\_dailymail''), general articles (``wikipedia''), science papers (``elsevier\_PHYS'') and patent documents (``big\_patent''); refer to Sec. \ref{sec:data_source} for more details about these datasets. The Assistant is given the text, and its ultimate goal is to provide a personalized summary, i.e.,one that aligns with all of the User's preferences. 

 To achieve this, the Assistant asks the User targeted questions aimed at understanding the User's summarization preferences, specifically a single question per conversation turn to which the User answers truthfully. After each turn, we request the Assistant to provide a summary of the User's text, using the information gathered (outside of the conversation). The Judge assesses whether the Assistant's summary correctly aligns with all of the User's summarization preferences. The maximum number of conversation turns is set to $10$.

\vspace{-0.2cm}
\section{Experiments}\label{sec:experiments}
\vspace{-0.1cm}
In this section, we present our experimental setup and results across the three benchmark tasks.  
Our experiments aim to answer three central questions:  
(1) How well can current LLMs uncover hidden information through multi-turn dialogue?  
(2) How efficiently do they adapt newly acquired knowledge to produce personalized responses?  
(3) How do these capabilities vary across reasoning-oriented and personalization-oriented tasks?

\begin{figure*}[t]
  \centering
      \includegraphics[width=1\linewidth]{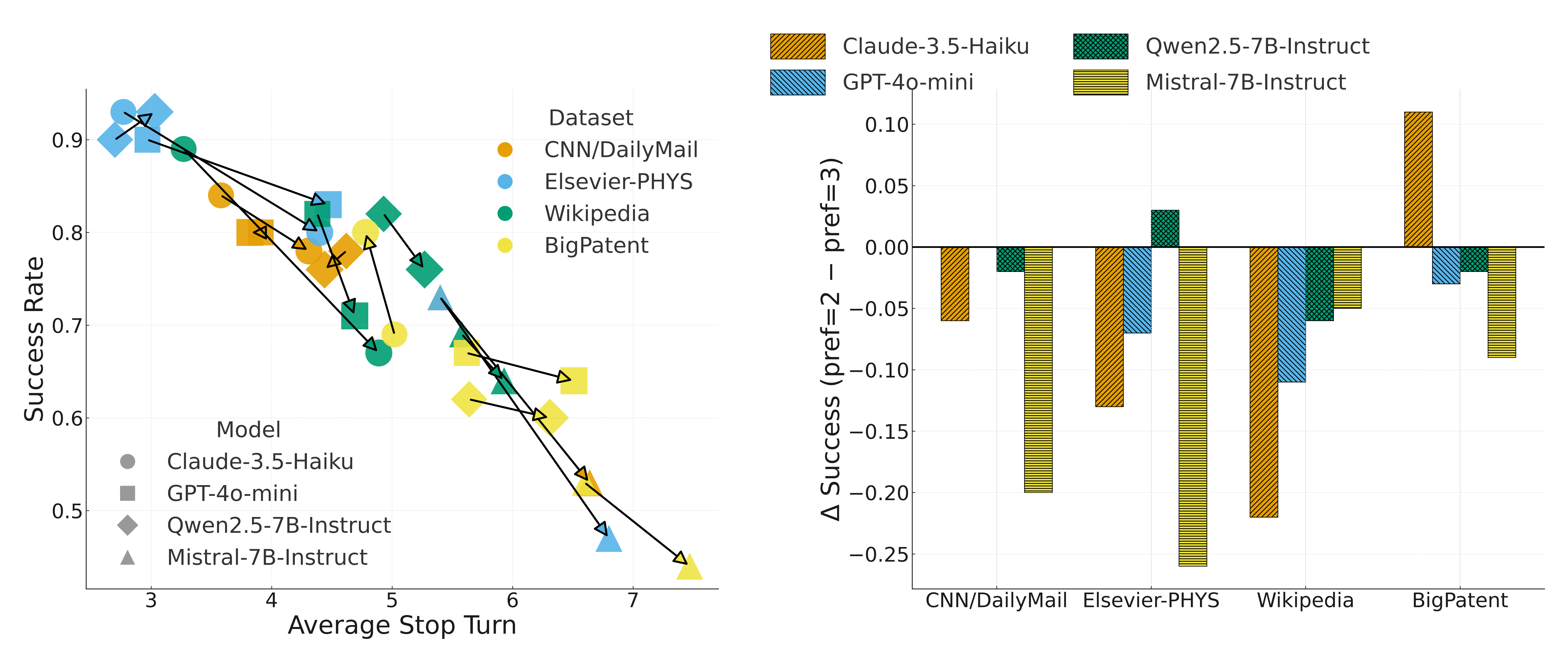}
  \caption{\textbf{Personalized Text Summarization results} across different models, topics, and preference settings. 
  (i) The scatter (left) reveals an efficiency–accuracy frontier: points nearer the upper-left indicate higher success with fewer turns; as the number of preferences increases (arrows), most (model, dataset) pairs shift down/right, showing longer elicitation and lower success. 
  (ii) Domain effects are pronounced—\textsc{Elsevier–PHYS} and \textsc{CNN/DailyMail} clusters tend to remain higher with shorter stops, while \textsc{BigPatent} drifts lower with longer stops, indicating greater difficulty in aligning to latent preferences for technical prose. 
  (iii) Model-wise, closed-source models typically occupy stronger regions, but open-weight \textsc{Qwen2.5-7B-Instruct} is often competitive in efficiency (shorter stop turns). 
  (iv) The difference-only bar chart (right) makes the penalty of increasing preference dimensionality explicit: most bars extend downward, quantifying the drop from \#Prefs=2 to \#Prefs=3 and highlighting which (topic, model) pairs degrade most.
  }
  \label{fig:pts_overview}
\end{figure*}

\vspace{-0.2cm}
\subsection{20 Questions Game}
\vspace{-0.1cm}
We present results in Fig.~\ref{fig:20q_combined}.  
All numbers are averaged over three runs; we have $20$ objects per topic.

\noindent \textbf{Main observations.} Performance drops sharply when the topic is unknown rather than disclosed, which is expected since the search space is much larger in the former case.  
The drop is less pronounced for the \emph{animal} topic because most Assistants open with a high-yield question (e.g., ``Is the object a living thing?''), quickly narrowing the hypothesis space.  
No single model dominates across all settings. Closed-source models (GPT-4o-mini, Claude-3.5-Haiku) tend to lead when the topic is known, while in the topic-unknown condition there are instances where open models (Qwen2.5-7B-Instruct, Mistral-7B-Instruct) perform best.  
Notably, Qwen2.5-7B-Instruct often identifies the correct object in fewer turns (see the \emph{landmark} topic), indicating stronger questioning efficiency even when overall accuracy is comparable. 

Although 20 Questions is not a personalization task, it isolates the latent-information \emph{discovery} process and thus offers a complementary view of models’ multi-turn reasoning: how well they plan, ask informative questions, and update hypotheses as evidence accumulates.  
We use these insights to interpret behavior on the personalization tasks.

\vspace{-0.2cm}
\subsection{Personalized Question Answering (PQA)}
\vspace{-0.2cm}
We present results in Fig.~\ref{fig:pqa_scatter_domains_models}.  
All numbers are averaged over three runs; each topic contains $10$ users.

\noindent \textbf{Main observations.} We observe substantial variability across topics: for a fixed model and preference count, success rates can differ by as much as $70\%$.  
For example, even with three preferences, all models achieve around $90\%$ success in \emph{Medical Care}, whereas performance drops to $20\%$–$50\%$ in \emph{Shopping Assistance}.  
Thus, PQA is not uniformly easy or hard; \emph{topic} strongly determines difficulty.  
Closed-source models (GPT-4o-mini, Claude-3.5-Haiku) are generally the top performers, though Mistral-7B-Instruct is competitive and occasionally surpasses them.  
Average stop-turn values remain modest in most settings— in only a third of the cases we have stop turn $\geq 6$—indicating that effective preference elicitation often occurs within the first few exchanges. An instance of two conversations for the Travel Restaurant topic can be found in Fig. \ref{fig:qa_example}; see also Fig. \ref{fig:qa_example2} in the Appendix.

\vspace{-0.2cm}
\subsection{Personalized Text Summarization (PTS)}
\vspace{-0.1cm}
We report results in Fig.~\ref{fig:pts_overview}.  
All numbers are averaged over three runs; each topic contains $15$ texts.

\textbf{Main observations.} Patterns broadly mirror those in PQA.  
Success rates decrease as the number of preferences increases, reflecting the added constraint burden.  
Closed-source models (GPT-4o-mini, Claude-3.5-Haiku) are, on average, strongest; however, Qwen2.5-7B-Instruct is frequently competitive and sometimes surpasses them on specific topics.  
Stop-turn values are generally modest—in most cases $<6$—suggesting that, for summarization, models can identify and apply key preferences with not too many clarification turns.

\vspace{-0.2cm}
\subsection{Error Type Analysis}\label{sec:error_analysis}
\vspace{-0.2cm}
To more comprehensively evaluate model behavior in personalization tasks, we conduct an \textbf{error type analysis} on the Personalized Question Answering task.  
The goal is not only to assess \emph{whether} the Assistant ultimately produces a correct answer, but also to examine \emph{how} its reasoning trajectory evolves and where it diverges from the intended path.  
We categorize observed errors into two groups: \emph{process errors}, which occur during the Assistant–User interaction, and \emph{result errors}, which appear in the Assistant’s final output.

\paragraph{Process Errors.}  
Process errors capture failures in the Assistant’s ability to discover and maintain the User’s latent preferences during conversation:
\begin{enumerate}
    \vspace{-2mm}
    \item \textbf{Preference Elicitation Error:} The Assistant fails to elicit one or more of the User’s preferences, often because its questions are too generic, redundant, or misdirected toward irrelevant aspects. \vspace{-3mm}
    \item \textbf{Preference Reinforcement Error:} The Assistant successfully uncovers a preference but later neglects or contradicts it in subsequent turns, indicating weak conversational memory or contextual tracking. \vspace{-2mm}
\end{enumerate}

\paragraph{Result Errors.}  
Result errors reflect issues in how the Assistant integrates the gathered preferences into its final recommendation:
\begin{enumerate}
    \vspace{-2mm}
    \item \textbf{Preference Omission Error:} The Assistant’s final answer disregards one or more explicitly identified preferences, reverting to a generic or incomplete response. \vspace{-3mm}
    \item \textbf{Preference Dilution Error:} The Assistant acknowledges the preferences but applies them only partially, producing responses that align loosely or inconsistently with the User’s intent. \vspace{-2mm}
\end{enumerate}

\vspace{-0.4cm}
Across models, the most frequent mistakes involve \emph{Preference Reinforcement} and \emph{Preference Dilution}, suggesting that LLMs often recognize preferences but fail to consistently preserve or operationalize them throughout multi-turn dialogue.  
These patterns highlight the need for improved mechanisms to represent and condition on user preferences during ongoing interaction.  

\vspace{-0.2cm}
\section{Conclusion}
\vspace{-0.2cm}
In this work we study the ability of LLMs to infer latent user preferences through conversation and to use them to generate personalized responses. To this end, we develop a benchmark consisting of three tasks. Our experiments show that in general LLMs possess the ability to uncover latent preferences. However, there are significant variations in the extent to which they demonstrate this ability, depending on the task and other factors. 
The next step is to explore approaches for improving the performance of LLMs on such tasks, particularly for weaker models and more challenging topics. Improvements should address not only success rates but also efficiency, by reducing the number of steps required to reach the correct answer.

\section*{Limitations}
We believe that our benchmark is well-rounded and provides a comprehensive view of LLMs’ ability to uncover latent user preferences in personalization tasks. However, it is subject to some limitations.  First, we tested only a limited number of models and topics in our experiments. This is because strong and accurate models (specifically GPT-4o) are required for both the Judge and the User agents in order to obtain reliable results, and these models are typically costly. As a result, we were able to run only a moderate number of experiments. Second, we do not provide any ways of improving the performance of LLMs, we only evaluate their inherent abilities of LLMs in unveiling latent user preferences in personalization tasks.

\bibliography{custom, anthology1}

\begin{thebibliography}{20}
\providecommand{\natexlab}[1]{#1}

\bibitem[{Afzoon et~al.(2024)Afzoon, Naseem, Beheshti, and Jamali}]{afzoon2024persobench}
Saleh Afzoon, Usman Naseem, Amin Beheshti, and Zahra Jamali. 2024.
\newblock Persobench: Benchmarking personalized response generation in large language models.
\newblock \emph{arXiv preprint arXiv:2410.03198}.

\bibitem[{Bai et~al.(2024)Bai, Liu, Bu, He, Liu, Zhou, Lin, Su, Ge, Zheng, and Ouyang}]{bai-etal-2024-mt}
Ge~Bai, Jie Liu, Xingyuan Bu, Yancheng He, Jiaheng Liu, Zhanhui Zhou, Zhuoran Lin, Wenbo Su, Tiezheng Ge, Bo~Zheng, and Wanli Ouyang. 2024.
\newblock \href {https://doi.org/10.18653/v1/2024.acl-long.401} {{MT}-bench-101: A fine-grained benchmark for evaluating large language models in multi-turn dialogues}.
\newblock In \emph{Proceedings of the 62nd Annual Meeting of the Association for Computational Linguistics (Volume 1: Long Papers)}, pages 7421--7454, Bangkok, Thailand. Association for Computational Linguistics.

\bibitem[{Busch et~al.(2025)Busch, Hoffmann, Rueger, van Dijk, Kader, Ortiz-Prado, Makowski, Saba, Hadamitzky, Kather et~al.}]{busch2025current}
Felix Busch, Lena Hoffmann, Christopher Rueger, Elon~HC van Dijk, Rawen Kader, Esteban Ortiz-Prado, Marcus~R Makowski, Luca Saba, Martin Hadamitzky, Jakob~Nikolas Kather, and 1 others. 2025.
\newblock Current applications and challenges in large language models for patient care: a systematic review.
\newblock \emph{Communications Medicine}, 5(1):26.

\bibitem[{Chu et~al.(2025)Chu, Wang, Xie, Zhu, Yan, Ye, Zhong, Hu, Liang, Yu et~al.}]{chu2025llm}
Zhendong Chu, Shen Wang, Jian Xie, Tinghui Zhu, Yibo Yan, Jinheng Ye, Aoxiao Zhong, Xuming Hu, Jing Liang, Philip~S Yu, and 1 others. 2025.
\newblock Llm agents for education: Advances and applications.
\newblock \emph{CoRR}.

\bibitem[{Hurst et~al.(2024)Hurst, Lerer, Goucher, Perelman, Ramesh, Clark, Ostrow, Welihinda, Hayes, Radford et~al.}]{hurst2024gpt}
Aaron Hurst, Adam Lerer, Adam~P Goucher, Adam Perelman, Aditya Ramesh, Aidan Clark, AJ~Ostrow, Akila Welihinda, Alan Hayes, Alec Radford, and 1 others. 2024.
\newblock Gpt-4o system card.
\newblock \emph{arXiv preprint arXiv:2410.21276}.

\bibitem[{Jiang et~al.(2023)Jiang, Liu, Liu, Zhao, Zhang, Gao, Zhang, Li, and Xiong}]{jiang2023clip}
Dongsheng Jiang, Yuchen Liu, Songlin Liu, Jin'e Zhao, Hao Zhang, Zhen Gao, Xiaopeng Zhang, Jin Li, and Hongkai Xiong. 2023.
\newblock From clip to dino: Visual encoders shout in multi-modal large language models.
\newblock \emph{arXiv preprint arXiv:2310.08825}.

\bibitem[{Jiang et~al.(2024)Jiang, Wang, Shen, Kim, and Kim}]{jiang2024survey}
Juyong Jiang, Fan Wang, Jiasi Shen, Sungju Kim, and Sunghun Kim. 2024.
\newblock A survey on large language models for code generation.
\newblock \emph{arXiv preprint arXiv:2406.00515}.

\bibitem[{Kumar et~al.(2024)Kumar, Viswanathan, Yerra, Salemi, Rossi, Dernoncourt, Deilamsalehy, Chen, Zhang, Agarwal et~al.}]{kumar2024longlamp}
Ishita Kumar, Snigdha Viswanathan, Sushrita Yerra, Alireza Salemi, Ryan~A Rossi, Franck Dernoncourt, Hanieh Deilamsalehy, Xiang Chen, Ruiyi Zhang, Shubham Agarwal, and 1 others. 2024.
\newblock Longlamp: A benchmark for personalized long-form text generation.
\newblock \emph{arXiv preprint arXiv:2407.11016}.

\bibitem[{Kwan et~al.(2024)Kwan, Zeng, Jiang, Wang, Li, Shang, Jiang, Liu, and Wong}]{kwan-etal-2024-mt}
Wai-Chung Kwan, Xingshan Zeng, Yuxin Jiang, Yufei Wang, Liangyou Li, Lifeng Shang, Xin Jiang, Qun Liu, and Kam-Fai Wong. 2024.
\newblock \href {https://doi.org/10.18653/v1/2024.emnlp-main.1124} {{MT}-eval: A multi-turn capabilities evaluation benchmark for large language models}.
\newblock In \emph{Proceedings of the 2024 Conference on Empirical Methods in Natural Language Processing}, pages 20153--20177, Miami, Florida, USA. Association for Computational Linguistics.

\bibitem[{Salemi et~al.(2024)Salemi, Mysore, Bendersky, and Zamani}]{salemi-etal-2024-lamp}
Alireza Salemi, Sheshera Mysore, Michael Bendersky, and Hamed Zamani. 2024.
\newblock \href {https://doi.org/10.18653/v1/2024.acl-long.399} {{L}a{MP}: When large language models meet personalization}.
\newblock In \emph{Proceedings of the 62nd Annual Meeting of the Association for Computational Linguistics (Volume 1: Long Papers)}, pages 7370--7392, Bangkok, Thailand. Association for Computational Linguistics.

\bibitem[{Tversky and Sattath(1979)}]{tversky1979preference}
Amos Tversky and Shmuel Sattath. 1979.
\newblock Preference trees.
\newblock \emph{Psychological review}, 86(6):542.

\bibitem[{Wang et~al.(2024)Wang, Du, Jiao, Lyu, Pang, Cui, Song, Wong, Shi, and Tu}]{wang-etal-2024-benchmarking}
Longyue Wang, Zefeng Du, Wenxiang Jiao, Chenyang Lyu, Jianhui Pang, Leyang Cui, Kaiqiang Song, Derek Wong, Shuming Shi, and Zhaopeng Tu. 2024.
\newblock \href {https://doi.org/10.18653/v1/2024.findings-acl.428} {Benchmarking and improving long-text translation with large language models}.
\newblock In \emph{Findings of the Association for Computational Linguistics: ACL 2024}, pages 7175--7187, Bangkok, Thailand. Association for Computational Linguistics.

\bibitem[{Wang et~al.(2025)Wang, Ma, Wang, Wu, Ji, Chen, Li, and Yuan}]{wang-etal-2025-survey}
Wenxuan Wang, Zizhan Ma, Zheng Wang, Chenghan Wu, Jiaming Ji, Wenting Chen, Xiang Li, and Yixuan Yuan. 2025.
\newblock \href {https://doi.org/10.18653/v1/2025.findings-acl.539} {A survey of {LLM}-based agents in medicine: How far are we from baymax?}
\newblock In \emph{Findings of the Association for Computational Linguistics: ACL 2025}, pages 10345--10359, Vienna, Austria. Association for Computational Linguistics.

\bibitem[{Xiao et~al.(2023)Xiao, Xu, Zhang, Wang, and Xia}]{xiao-etal-2023-evaluating}
Changrong Xiao, Sean~Xin Xu, Kunpeng Zhang, Yufang Wang, and Lei Xia. 2023.
\newblock \href {https://doi.org/10.18653/v1/2023.bea-1.52} {Evaluating reading comprehension exercises generated by {LLM}s: A showcase of {C}hat{GPT} in education applications}.
\newblock In \emph{Proceedings of the 18th Workshop on Innovative Use of NLP for Building Educational Applications (BEA 2023)}, pages 610--625, Toronto, Canada. Association for Computational Linguistics.

\bibitem[{Yang et~al.(2025)Yang, Li, Yang, Zhang, Hui, Zheng, Yu, Gao, Huang, Lv et~al.}]{yang2025qwen3}
An~Yang, Anfeng Li, Baosong Yang, Beichen Zhang, Binyuan Hui, Bo~Zheng, Bowen Yu, Chang Gao, Chengen Huang, Chenxu Lv, and 1 others. 2025.
\newblock Qwen3 technical report.
\newblock \emph{arXiv preprint arXiv:2505.09388}.

\bibitem[{Zan et~al.(2023)Zan, Chen, Zhang, Lu, Wu, Guan, Yongji, and Lou}]{zan-etal-2023-large}
Daoguang Zan, Bei Chen, Fengji Zhang, Dianjie Lu, Bingchao Wu, Bei Guan, Wang Yongji, and Jian-Guang Lou. 2023.
\newblock \href {https://doi.org/10.18653/v1/2023.acl-long.411} {Large language models meet {NL}2{C}ode: A survey}.
\newblock In \emph{Proceedings of the 61st Annual Meeting of the Association for Computational Linguistics (Volume 1: Long Papers)}, pages 7443--7464, Toronto, Canada. Association for Computational Linguistics.

\bibitem[{Zhang et~al.(2024)Zhang, Lu, and Jaitly}]{zhang-etal-2024-probing}
Yizhe Zhang, Jiarui Lu, and Navdeep Jaitly. 2024.
\newblock \href {https://doi.org/10.18653/v1/2024.acl-long.82} {Probing the multi-turn planning capabilities of {LLM}s via 20 question games}.
\newblock In \emph{Proceedings of the 62nd Annual Meeting of the Association for Computational Linguistics (Volume 1: Long Papers)}, pages 1495--1516, Bangkok, Thailand. Association for Computational Linguistics.

\bibitem[{Zhao et~al.(2025{\natexlab{a}})Zhao, Hong, Liu, Hazarika, and Lin}]{zhaollms}
Siyan Zhao, Mingyi Hong, Yang Liu, Devamanyu Hazarika, and Kaixiang Lin. 2025{\natexlab{a}}.
\newblock Do llms recognize your preferences? evaluating personalized preference following in llms.
\newblock In \emph{The Thirteenth International Conference on Learning Representations}.

\bibitem[{Zhao et~al.(2025{\natexlab{b}})Zhao, Vania, Kayal, Khan, Cohen, and Yilmaz}]{zhao-etal-2025-personalens}
Zheng Zhao, Clara Vania, Subhradeep Kayal, Naila Khan, Shay~B Cohen, and Emine Yilmaz. 2025{\natexlab{b}}.
\newblock \href {https://doi.org/10.18653/v1/2025.findings-acl.927} {{P}ersona{L}ens: A benchmark for personalization evaluation in conversational {AI} assistants}.
\newblock In \emph{Findings of the Association for Computational Linguistics: ACL 2025}, pages 18023--18055, Vienna, Austria. Association for Computational Linguistics.

\bibitem[{Zhu et~al.(2024)Zhu, Liu, Dong, Xu, Huang, Kong, Chen, and Li}]{zhu-etal-2024-multilingual}
Wenhao Zhu, Hongyi Liu, Qingxiu Dong, Jingjing Xu, Shujian Huang, Lingpeng Kong, Jiajun Chen, and Lei Li. 2024.
\newblock \href {https://doi.org/10.18653/v1/2024.findings-naacl.176} {Multilingual machine translation with large language models: Empirical results and analysis}.
\newblock In \emph{Findings of the Association for Computational Linguistics: NAACL 2024}, pages 2765--2781, Mexico City, Mexico. Association for Computational Linguistics.

\end{thebibliography}

\newpage
\appendix

\section{Additional Experiments}

\subsection{Active vs Passive User Comparison}

\begin{table*}[t]
\centering
\small
\begin{tabular}{cccccccc}
\hline
\textbf{Topic}& \textbf{Model}& \multicolumn{2}{c}{\textbf{Passive User}} & \multicolumn{2}{c}{\textbf{Active User}}  \\
 &  & \textbf{Success Rate} & \textbf{Stop Turn} & \textbf{Success Rate} & \textbf{Stop Turn} \\
\hline
\rowcolor{gr}
 \cellcolor{gr2}& Claude-3.5-Haiku & 0.78 ± 0.06 & 4.31 ± 0.41 & 0.98 ± 0.03 & 4.31 ± 0.30 \\
 \rowcolor{bl}
 \cellcolor{gr2}\multirow{-2}{*}{\shortstack{CNN\\DailyMail}}& GPT-4o-mini & 0.80 ± 0.05 & 3.82 ± 0.37 & 1.00 ± 0.00 & 4.49 ± 0.38\\
\hline
\rowcolor{gr}
 \cellcolor{bl2}& Claude-3.5-Haiku & 0.67 ± 0.05 & 4.89 ± 0.80 & 0.80 ± 0.05 & 5.69 ± 0.47 \\
 \rowcolor{bl}
 \cellcolor{bl2}\multirow{-2}{*}{\shortstack{Wikipedia}} & GPT-4o-mini & 0.71 ± 0.03 & 4.69 ± 0.32 & 0.98 ± 0.03 & 5.33 ± 0.83 \\
\hline
\rowcolor{gr}
 \cellcolor{gr2}& Claude-3.5-Haiku & 0.80 ± 0.00  & 4.78 ± 0.08 & 0.96 ± 0.06 & 4.53 ± 0.73 \\
 \rowcolor{bl}
 \cellcolor{gr2}\multirow{-2}{*}{\shortstack{Big \\ Patent}} & GPT-4o-mini & 0.64 ± 0.08 & 6.51 ± 0.60 & 1.00 ± 0.00 & 4.60 ± 0.18 \\
\hline
\end{tabular}
\caption{Comparison of the performance of a passive (default case) and an active user on the Text Summarization task.}
\label{tab:pass_act}
\end{table*}

Our default benchmark assumes a \emph{passive} User, defined as someone who only responds to the Assistant’s questions without volunteering additional information unless explicitly prompted (see Sec.~\ref{sec:intro}).  
This assumption reflects the common real-world scenario in which users provide short, underspecified prompts and expect the model to take initiative in eliciting what is needed. Benchmarking under this assumption provide an assessment of the {\it worst case}, yet still common, scenario. 

To explore how this assumption affects task difficulty, we consider a variant of the Personalized Text Summarization task with an \emph{active} User.  
In this setting, the Assistant still poses a question each turn but also provides a preliminary summary.  
The User may either answer the question or proactively comment on the summary (e.g. by requesting a more concise summary or requesting the inclusion of specific information from the original text), whichever they judge more helpful for guiding the Assistant toward alignment. Note that in this case, since users are actively engaging the conversation, we expect that they will be directly revealing their preferences, therefore we expect that the Assistant will be able to achieve a much higher success rate. See Fig. \ref{fig:pass_act} for an example conversation with a passive and an active User.

\begin{figure*}[ht]
\centering
{\small
    \begin{minipage}[t]{1\textwidth}
        \begin{tcolorbox}[colback=black!5, colframe=gray!50, boxrule=0pt, arc=0mm]
        \vspace{-0.2cm}
        \textbf{Preferences} \\
        1. Focus primarily on the medical details of Theia's injuries and treatment rather than the emotional aspects of her story.\\
        2. Emphasize the fundraising efforts and financial aspects of her care more than the sequence of traumatic events.\\
        3. Give equal attention to the similar story from Tampa, FL as a parallel case rather than treating it as a brief footnote.
        \vspace{-0.2cm}
        \end{tcolorbox}
    \end{minipage}
    \\
    \begin{minipage}[t]{1\textwidth}
        \begin{tcolorbox}[colback=green!5, colframe=gray!50, boxrule=0pt, arc=0mm]
        \textbf{Text:} (CNN) Never mind cats having nine lives. A stray pooch in Washington State has used up at least three of her own after being hit by a car, apparently whacked on the head with a hammer in a misguided mercy killing and then buried in a field -- only to survive. (...) 
        She suffered a dislocated jaw, leg injuries and a caved-in sinus cavity -- and still requires surgery to help her breathe. The veterinary hospital's Good Samaritan Fund committee awarded some money to help pay for the dog's treatment, but Mellado has set up a fundraising page to help meet the remaining cost of the dog's care. (...) 
        Theia is not the only animal to apparently rise from the grave in recent weeks. A cat in Tampa, Florida, found seemingly dead after he was hit by a car in January (...)
        \end{tcolorbox}
    \end{minipage}
    \\
    \begin{minipage}[t]{1\textwidth}
        \begin{tcolorbox}[colback=blue!5, colframe=gray!50, boxrule=0pt, arc=2mm]
        \vspace{-0.2cm}
        \textbf{Passive User} \\
        1.A: {\blue Would you prefer a summary that focuses more on the dog's survival story or the broader context of animal resilience?} \\
        1.U: {\red I would prefer a summary that focuses more on the medical details of the dog's injuries and treatment.}\\
        2.A: {\blue Would you like the medical details presented in a clinical, technical tone or in a more narrative, empathetic style?}\\
        2.U: {\red I would like the medical details presented in a clinical, technical tone.} \\
        ...... 
        \vspace{-0.2cm}
        \end{tcolorbox}
    \end{minipage}
    \\
    \begin{minipage}[t]{1\textwidth}
        \begin{tcolorbox}[colback=gray!5, colframe=gray!50, boxrule=0pt, arc=2mm]
        \vspace{-0.2cm}
        \textbf{Active User}\\
        1.A: {\blue **Summary**:
        A resilient stray dog named Theia survived multiple life-threatening incidents, including being hit by a car, apparently struck with a hammer, and buried in a field. Four days after her presumed death, she crawled to a nearby farm and was rescued by Sara Mellado. Despite severe injuries like a dislocated jaw and damaged sinus cavity, Theia is receiving medical treatment at Washington State University's Veterinary Teaching Hospital. A fundraising effort has already exceeded \$10,000 to support her recovery, with Mellado committed to finding her a loving, safe home.
        \\
        **Question**:
        Would you prefer a summary that focuses more on the emotional aspects of Theia's survival story, or one that emphasizes the factual details of her medical condition and rescue?} \\
        1.U: {\red I would prefer a summary that emphasizes the factual details of Theia's medical condition and the treatment she is receiving.}\\
        2.A: {\blue **Summary**:
        Theia, a one-year-old stray dog, suffered extensive medical trauma including a car accident, potential blunt force injury, and burial. Her specific medical conditions include a dislocated jaw, leg injuries, and a caved-in sinus cavity requiring surgical intervention. Currently receiving treatment at Washington State University's Veterinary Teaching Hospital, her care is partially funded by the Good Samaritan Fund. Sara Mellado has established additional fundraising, which has already surpassed the \$10,000 target, to cover her extensive medical procedures necessary for nasal damage repair and jaw reset.
        \\
        **Question**:
        Would you like me to include more specific details about the types of medical procedures Theia requires or the potential long-term prognosis for her recovery?
        }\\
        2.U: {\red Please include more details on the fundraising efforts and financial aspects of Theia's care.}\\
        ......
        \vspace{-0.2cm}
        \end{tcolorbox} 
    \end{minipage}
}
\vspace{-0.7cm}
\caption{An instance of the Personalized Text Summarization task for the ``CNN DailyMail'' topic. We present two conversations illustrating cases where the User agent acts as either a passive or an active participant. A passive User only responds to the Assistant’s queries, whereas an active User either provides feedback on the summary or answers the question. Indeed, we observe that the active User answers the Assistant’s question in the first turn, while in the second turn it ignores the question and instead provides direct feedback on the summary, requesting changes to better align with the latent preferences.}
\label{fig:pass_act}
\end{figure*}
Table~\ref{tab:pass_act} shows that success rates with an active User are dramatically higher, often approaching $100\%$.  
While this result demonstrates that most models can produce high-quality personalized outputs once preferences are explicitly surfaced, it also underscores that such an interaction style is largely impractical in ordinary use.  
Real users rarely maintain that level of engagement or meta-awareness about their own preferences.  
Hence, the \emph{passive} setting—where the model must take the lead in discovering latent information—remains the more realistic and challenging benchmark for studying personalization in LLMs.

\subsection{Judge Evaluation}\label{sec:judge_eval}

One key component of the benchmark is the Judge agent, whose role is to assess whether the Assistant’s output satisfies all of the User’s latent preferences. It is clear that the Judge plays a critical role in ensuring accurate and reliable results. Therefore, selecting a strong and reliable LLM for the Judge agent is crucial. In our experiments, we employed GPT-4o, given its well-established capabilities. To further justify this choice, we conducted additional experiments to assess its suitability.

More precisely, we evaluate the extent to which the Judge LLM follows the instructions provided to it in the Personalized Question Answering task. To do so, we employ three competent LLMs, Claude 3.7 Sonnet, Claude 4 Sonnet, and Claude 4.5 Sonnet, to assess instruction adherence. Each model receives as input the User-Assistant conversation, the Assistant’s responses to the User’s questions after each turn, and the Judge’s assessments of those responses. The models then determine whether the Judge followed its instructions, and the final assessment is computed using majority voting. We report the percentage of conversations in which the Judge adheres to the given instructions. 

We test three LLM models as Judges, GPT-4o (our choice), GPT-4o-mini, and Mistral-7B-Instruct. The results are presented in Table \ref{tab:judge_eval}. GPT-4o clearly outperforms the other models in terms of instruction adherence, thereby validating our choice to use it as the Judge model.

\begin{table*}[t]
\centering
\small
\begin{tabular}{cc}
\hline
\textbf{Model} & \textbf{Instruction Adherence Rate}  \\
\hline 
\rowcolor{gr} GPT-4o & 0.74 \\
\hline
\rowcolor{bl} GPT-4o-mini & 0.41 \\
\hline
\rowcolor{gr} Mistral-7B-Instruct & 0.11 \\
\hline
\end{tabular}
\caption{The results of evaluating different LLMs in the role of the Judge. The instruction adherence rate is the percentage of conversations in which the model adheres to the Judge's instructions. GPT-4o clearly outperforms the other models in terms of instruction adherence.}
\label{tab:judge_eval}
\end{table*}

\section{Additional Experimental Details}

\subsection{Experiment Tables}

For completeness, we include the tables of results from the experiments discussed in the main text. The results for the 20 Question Game task are given in Table \ref{tab:20q}, for the Personalized Question Answering task in Table \ref{tab:person_qa}, and for the Text Summarization task in Table \ref{tab:text_summ}.

\begin{table*}[t]
\centering
\small
\begin{tabular}{cccccc}
\hline
\textbf{Topic} & \textbf{Model} & \multicolumn{2}{c}{\textbf{Topic Known}} & \multicolumn{2}{c}{\textbf{Topic Unknown}} \\
\multicolumn{2}{c}{} &  \textbf{Success Rate} & \textbf{Stop Turn} & \textbf{Success Rate} & \textbf{Stop Turn} \\
\hline
\rowcolor{gr}
\cellcolor{gr2} & Claude-3.5-Haiku & 0.90 $\pm$ 0.04 & 7.68 ± 0.73 & 0.83 $\pm$ 0.02 & 11.35 ± 0.61 \\
\rowcolor{bl}
\cellcolor{gr2} & GPT-4o-mini & 0.93 $\pm$ 0.02 & 6.60 ± 0.43 & 0.97 $\pm$ 0.02 & 8.25 ± 0.52 \\
\rowcolor{gr}
\cellcolor{gr2} & Qwen2.5-7B-Instruct & 0.75 $\pm$ 0.00 & 9.27 ± 1.64 & 0.58 $\pm$ 0.08 & 12.55 ± 1.69 \\
\rowcolor{bl}
\cellcolor{gr2} \multirow{-4}{*}{Animal} & Mistral-7B-Instruct & 0.83 $\pm$ 0.08 & 9.02 ± 2.18 & 0.78 $\pm$ 0.02 & 11.05 ± 0.35 \\
\hline
 \rowcolor{gr}
 \cellcolor{bl2}& Claude-3.5-Haiku & 0.90 $\pm$ 0.07 & 9.78 ± 0.58 & 0.07 $\pm$ 0.02 & 19.70 ± 0.13 \\
 \rowcolor{bl}
 \cellcolor{bl2}& GPT-4o-mini & 0.93 $\pm$ 0.02 & 8.35 ± 0.66 & 0.20 $\pm$ 0.04 & 19.28 ± 0.30 \\
 \rowcolor{gr}
 \cellcolor{bl2}& Qwen2.5-7B-Instruct & 0.75 $\pm$ 0.00 & 9.72 ± 0.80 & 0.40 $\pm$ 0.14 & 15.08 ± 2.56 \\
 \rowcolor{bl}
 \cellcolor{bl2}\multirow{-4}{*}{Fruit} & Mistral-7B-Instruct & 0.65 $\pm$ 0.04 & 11.75 ± 0.44 & 0.45 $\pm$ 0.15 & 16.32 ± 0.94 \\
\hline
 \rowcolor{gr}
\cellcolor{gr2}& Claude-3.5-Haiku & 0.93 $\pm$ 0.02 & 7.38 ± 0.49 & 0.18 $\pm$ 0.05 & 19.37 ± 0.28 \\
\rowcolor{bl}
 \cellcolor{gr2}& GPT-4o-mini & 0.93 $\pm$ 0.06 & 8.00 ± 0.46 & 0.13 $\pm$ 0.02 & 19.65 ± 0.10 \\
 \rowcolor{gr}
 \cellcolor{gr2}& Qwen2.5-7B-Instruct & 0.78 $\pm$ 0.02 & 7.52 ± 0.29 & 0.47 $\pm$ 0.10 & 13.02 ± 1.90 \\
 \rowcolor{bl}
\cellcolor{gr2}\multirow{-4}{*}{Landmark} & Mistral-7B-Instruct & 0.80 $\pm$ 0.07 & 9.38 ± 1.44 & 0.20 $\pm$ 0.07 & 18.37 ± 0.53 \\
\hline
 \rowcolor{gr}
 \cellcolor{bl2}& Claude-3.5-Haiku & 0.95 $\pm$ 0.04 & 7.50 ± 0.31 & 0.30 $\pm$ 0.00 & 18.62 ± 0.08\\
 \rowcolor{bl}
 \cellcolor{bl2}& GPT-4o-mini & 0.75 $\pm$ 0.14 & 13.07 ± 1.03 & 0.23 $\pm$ 0.08 & 18.92 ± 0.39\\
 \rowcolor{gr}
 \cellcolor{bl2}& Qwen2.5-7B-Instruct & 0.62 $\pm$ 0.12 & 10.57 ± 2.28 & 0.40 $\pm$ 0.07 & 15.43 ± 1.10 \\
 \rowcolor{bl}
 \cellcolor{bl2}\multirow{-4}{*}{Sport} & Mistral-7B-Instruct & 0.80 $\pm$ 0.07 & 10.68 ± 2.09 & 0.07 $\pm$ 0.02 & 19.55 ± 0.25\\
\hline
 \rowcolor{gr}
 \cellcolor{gr2}& Claude-3.5-Haiku & 0.92 $\pm$ 0.02 & 8.38 ± 1.05 & 0.55 $\pm$ 0.04 & 16.10 ± 1.18\\
 \rowcolor{bl}
 \cellcolor{gr2}& GPT-4o-mini & 0.95 $\pm$ 0.00 & 7.80 ± 0.64 & 0.57 $\pm$ 0.08 & 16.95 ± 0.13 \\
 \rowcolor{gr}
 \cellcolor{gr2}& Qwen2.5-7B-Instruct& 0.65 $\pm$ 0.11 & 10.70 ± 2.36 & 0.52 $\pm$ 0.10 & 14.07 ± 1.59 \\
 \rowcolor{bl}
 \cellcolor{gr2}\multirow{-4}{*}{Weather} & Mistral-7B-Instruct & 0.68 $\pm$ 0.12 & 12.12 ± 2.52 & 0.12 $\pm$ 0.05 & 19.35 ± 0.13 \\
\hline
\end{tabular}
\caption{Results of the ``20 Questions Game'' experiments across different models, topics, and conditions (topic known vs unknown).}
\label{tab:20q}
\end{table*}
\begin{table*}[t]
\centering
\small
\begin{tabular}{cccccccc}
\hline
\textbf{Topic}& \textbf{Model}& \multicolumn{2}{c}{\textbf{\# Preferences=1}} & \multicolumn{2}{c}{\textbf{\# Preferences=2}} & \multicolumn{2}{c}{\textbf{\# Preferences=3}}  \\
 &  & \textbf{Success Rate} & \textbf{Stop Turn} & \textbf{Success Rate} & \textbf{Stop Turn} & \textbf{Success Rate} & \textbf{Stop Turn} \\
\hline
\rowcolor{gr}
 \cellcolor{gr2}& Claude-3.5-Haiku & 1.00 ± 0.00 & 2.43 ± 0.33 & 0.97 ± 0.05 & 3.43 ± 0.06 & 0.90 ± 0.00 & 4.40 ± 0.35 \\
 \rowcolor{bl}
 \cellcolor{gr2}& GPT-4o-mini & 1.00 ± 0.00 & 1.77 ± 0.41 & 0.97 ± 0.05 & 2.53 ± 0.81 & 0.93 ± 0.05 & 3.60 ± 0.75 \\
 \rowcolor{gr}
 \cellcolor{gr2}& Qwen2.5-7B-Instruct & 0.97 ± 0.05 & 2.40 ± 0.56 & 0.93 ± 0.05 & 3.10 ± 0.26 & 0.87 ± 0.12 & 5.17 ± 1.24 \\
 \rowcolor{bl}
 \cellcolor{gr2}\multirow{-4}{*}{\shortstack{Medical\\Care}}& Mistral-7B-Instruct & 0.93 ± 0.05 & 1.93 ± 0.57 & 1.00 ± 0.00 & 2.53 ± 0.21 & 0.93 ± 0.05 & 2.30 ± 0.44 \\
\hline
\rowcolor{gr}
 \cellcolor{bl2}& Claude-3.5-Haiku & 0.87 ± 0.05 & 3.97 ± 0.31 & 0.83 ± 0.09 & 5.60 ± 0.89 & 0.83 ± 0.05 & 5.83 ± 0.67 \\
 \rowcolor{bl}
 \cellcolor{bl2}& GPT-4o-mini & 0.83 ± 0.05 & 4.30 ± 0.20 & 0.67 ± 0.05 & 6.43 ± 0.42 & 0.87 ± 0.05 & 5.57 ± 0.49\\
 \rowcolor{gr}
 \cellcolor{bl2}& Qwen2.5-7B-Instruct & 0.83 ± 0.05 & 4.27 ± 0.86 & 0.77 ± 0.12 & 5.93 ± 0.74 & 0.70 ± 0.08 & 6.63 ± 0.23\\
 \rowcolor{bl}
 \cellcolor{bl2}\multirow{-4}{*}{\shortstack{Personal\\Finance}} & Mistral-7B-Instruct & 0.80 ± 0.08 & 4.30 ± 0.40 & 0.77 ± 0.05 & 4.77 ± 0.61 & 0.83 ± 0.12 & 4.37 ± 0.23\\
\hline
\rowcolor{gr}
 \cellcolor{gr2}& Claude-3.5-Haiku & 0.93 ± 0.05 & 3.10 ± 0.50 & 0.73 ± 0.05 & 5.03 ± 0.47 & 0.73 ± 0.05 & 4.80 ± 0.10\\
 \rowcolor{bl}
 \cellcolor{gr2}& GPT-4o-mini & 0.87 ± 0.09 & 4.60 ± 0.61 & 0.73 ± 0.09 & 6.47 ± 0.78 & 0.70 ± 0.00 & 5.63 ± 0.06 \\
 \rowcolor{gr}
 \cellcolor{gr2}& Qwen2.5-7B-Instruct & 0.80 ± 0.08 & 4.77 ± 0.90 & 0.70 ± 0.00 & 6.00 ± 1.04 & 0.67 ± 0.09 & 6.00 ± 0.26 \\
 \rowcolor{bl}
 \cellcolor{gr2}\multirow{-4}{*}{\shortstack{Pet\\Ownership}} & Mistral-7B-Instruct & 0.83 ± 0.05 & 3.93 ± 0.74 & 0.87 ± 0.05 & 5.60 ± 0.72 & 0.67 ± 0.12 & 5.30 ± 1.04\\
\hline
\rowcolor{gr}
 \cellcolor{bl2}& Claude-3.5-Haiku & 0.60 ± 0.08 & 6.20 ± 0.30 & 0.33 ± 0.05 & 7.93 ± 0.57 & 0.50 ± 0.08 & 7.57 ± 0.31 \\
 \rowcolor{bl}
 \cellcolor{bl2}& GPT-4o-mini & 0.63 ± 0.17 & 6.23 ± 0.67 & 0.43 ± 0.05 & 7.67 ± 0.35 & 0.27 ± 0.05 & 8.50 ± 0.44\\
 \rowcolor{gr}
 \cellcolor{bl2}& Qwen2.5-7B-Instruct & 0.53 ± 0.09 & 6.73 ± 0.71 & 0.40 ± 0.00 & 8.57 ± 0.61 & 0.17 ± 0.05 & 9.07 ± 0.45 \\
 \rowcolor{bl}
 \cellcolor{bl2}\multirow{-4}{*}{\shortstack{Shopping\\Assistance}} & Mistral-7B-Instruct & 0.73 ± 0.05 & 5.43 ± 0.42 & 0.53 ± 0.05 & 7.33 ± 0.38 & 0.33 ± 0.05 & 8.40 ± 0.36\\
\hline
\rowcolor{gr}
 \cellcolor{gr2}& Claude-3.5-Haiku & 0.87 ± 0.05 & 4.07 ± 0.42 & 0.67 ± 0.09 & 5.23 ± 0.57 & 0.67 ± 0.12 & 6.13 ± 0.29 \\
 \rowcolor{bl}
 \cellcolor{gr2}& GPT-4o-mini & 0.90 ± 0.08 & 2.77 ± 0.61 & 0.40 ± 0.00 & 7.53 ± 0.45 & 0.40 ± 0.08 & 7.40 ± 0.72 \\
 \rowcolor{gr}
 \cellcolor{gr2}& Qwen2.5-7B-Instruct & 0.77 ± 0.09 & 4.20 ± 0.79 & 0.37 ± 0.05 & 7.80 ± 0.46 & 0.20 ± 0.08 & 8.80 ± 0.30 \\
 \rowcolor{bl}
 \cellcolor{gr2}\multirow{-4}{*}{\shortstack{Travel\\Restaurant}} & Mistral-7B-Instruct & 0.77 ± 0.09 & 4.73 ± 0.71 & 0.30 ± 0.00 & 7.73 ± 0.29 & 0.30 ± 0.00 & 8.03 ± 0.21 \\
\hline
\end{tabular}
\caption{Results of the Personalized Question Answering experiments across different models, topics, and preference settings.}
\label{tab:person_qa}
\end{table*}
\begin{table*}[t]
\centering
\small
\begin{tabular}{cccccccc}
\hline
\textbf{Topic}& \textbf{Model}& \multicolumn{2}{c}{\textbf{\# Preferences=2}} & \multicolumn{2}{c}{\textbf{\# Preferences=3}}  \\
 &  & \textbf{Success Rate} & \textbf{Stop Turn} & \textbf{Success Rate} & \textbf{Stop Turn} \\
\hline
\rowcolor{gr}
 \cellcolor{gr2}& Claude-3.5-Haiku & 0.84 ± 0.03 & 3.58 ± 0.04 & 0.78 ± 0.06 & 4.31 ± 0.41 \\
 \rowcolor{bl}
 \cellcolor{gr2}& GPT-4o-mini & 0.80 ± 0.00 & 3.91 ± 0.10 & 0.80 ± 0.05 & 3.82 ± 0.37 \\
 \rowcolor{gr}
 \cellcolor{gr2}& Qwen2.5-7B-Instruct & 0.78 ± 0.03 & 4.62 ± 0.51 & 0.76 ± 0.03 & 4.44 ± 0.37 \\
 \rowcolor{bl}
 \cellcolor{gr2}\multirow{-4}{*}{\shortstack{CNN\\DailyMail}}& Mistral-7B-Instruct & 0.73 ± 0.05 & 5.40 ± 0.81 & 0.53 ± 0.09 & 6.64 ± 0.44 \\
\hline
\rowcolor{gr}
 \cellcolor{bl2}& Claude-3.5-Haiku & 0.93 ± 0.05 & 2.77 ± 0.40 & 0.80 ± 0.00 & 4.40 ± 0.44 \\
 \rowcolor{bl}
 \cellcolor{bl2}& GPT-4o-mini & 0.90 ± 0.08 & 2.97 ± 0.47 & 0.83 ± 0.05 & 4.47 ± 0.35 \\
 \rowcolor{gr}
 \cellcolor{bl2}& Qwen2.5-7B-Instruct & 0.90 ± 0.08 & 2.70 ± 0.50 & 0.93 ± 0.05 & 3.03 ± 0.40 \\
 \rowcolor{bl}
 \cellcolor{bl2}\multirow{-4}{*}{\shortstack{Elsevier\\PHYS}} & Mistral-7B-Instruct & 0.73 ± 0.05 & 5.40 ± 0.10 & 0.47 ± 0.09 & 6.80 ± 1.10 \\
\hline
\rowcolor{gr}
 \cellcolor{gr2}& Claude-3.5-Haiku & 0.89 ± 0.03 & 3.27 ± 0.59 & 0.67 ± 0.05 & 4.89 ± 0.80 \\
 \rowcolor{bl}
 \cellcolor{gr2}& GPT-4o-mini & 0.82 ± 0.06 & 4.38 ± 0.08 & 0.71 ± 0.03 & 4.69 ± 0.32 \\
 \rowcolor{gr}
 \cellcolor{gr2}& Qwen2.5-7B-Instruct & 0.82 ± 0.08 & 4.93 ± 0.77 & 0.76 ± 0.03 & 5.27 ± 0.44 \\
 \rowcolor{bl}
 \cellcolor{gr2}\multirow{-4}{*}{\shortstack{Wikipedia}} & Mistral-7B-Instruct & 0.69 ± 0.06 & 5.58 ± 0.44 & 0.64 ± 0.03 & 5.93 ± 0.57 \\
\hline
\rowcolor{gr}
 \cellcolor{bl2}& Claude-3.5-Haiku & 0.69 ± 0.03 & 5.02 ± 0.27 & 0.80 ± 0.00  & 4.78 ± 0.08 \\
 \rowcolor{bl}
 \cellcolor{bl2}& GPT-4o-mini & 0.67 ± 0.05 & 5.62 ± 0.45 & 0.64 ± 0.08 & 6.51 ± 0.60 \\
 \rowcolor{gr}
 \cellcolor{bl2}& Qwen2.5-7B-Instruct & 0.62 ± 0.08 & 5.64 ± 0.39 & 0.60 ± 0.00 & 6.31 ± 0.14 \\
 \rowcolor{bl}
 \cellcolor{bl2}\multirow{-4}{*}{\shortstack{Big \\ Patent}} & Mistral-7B-Instruct & 0.53 ± 0.05 & 6.60 ± 0.35 & 0.44 ± 0.06 & 7.47 ± 0.41 \\
\hline
\end{tabular}
\caption{Results of the Text Summarization experiments across different models, topics, and preference settings. 
}
\label{tab:text_summ}
\end{table*}

\subsection{Sample Conversations}

In Figure \ref{fig:qa_example2} we present an instance of the Personalized Question Answering task for the ``Travel Restaurant topic'', along with the conversations of two models. This is the same instance discussed in Fig \ref{fig:qa_example} in the main text. Here we complete this example by providing the conversations conducted by the other two models: Claude-3.5-Haiku and Mistral-7B-Instruct. The observations are similar to the ones from the main text. The model that performs best in this case, Claude-3.5-Haiku, begins with general questions, allowing it to identify the latent preference (e.g., avoid spicy food) earlier than the other model. In contrast, Mistral-7B-Instruct immediately narrows its focus after the first turn (e.g., asking about specific dishes) and becomes 'trapped' in a narrow line of questioning for two turns.

\begin{figure*}[ht]
\centering
{\small
    \begin{minipage}[t]{0.34\textwidth}
        \begin{tcolorbox}[colback=black!5, colframe=gray!50, boxrule=0pt, arc=0mm]
        \vspace{-0.5cm}
        \textbf{Preference:} I avoid spicy foods. \vspace{-0.2cm}
        \end{tcolorbox}
    \end{minipage}
    \begin{minipage}[t]{0.65\textwidth}
        \begin{tcolorbox}[colback=green!5, colframe=gray!50, boxrule=0pt, arc=0mm]
        \vspace{-0.5cm}
        \textbf{Q:} Can you recommend some restaurants to try during my visit to Mumbai?\vspace{-0.2cm}
        \end{tcolorbox}
    \end{minipage}
    \\
    \begin{minipage}[t]{1\textwidth}
        \begin{tcolorbox}[colback=blue!5, colframe=gray!50, boxrule=0pt, arc=2mm]
        \vspace{-0.2cm}
        \textbf{Claude-3.5-Haiku}\\
        1.A: {\blue What type of dining experience are you looking for during your visit to Mumbai?}\\
        1.U: {\red I'm looking for a variety of dining experiences, from casual to fine dining, that showcase the diverse culinary scene in Mumbai.}\\
        2.A: {\blue Are there any specific cuisines or types of food you're particularly interested in trying during your Mumbai trip?}\\
        2.U: {\red I'm interested in trying a mix of local Indian cuisine and international flavors to really get a taste of what Mumbai has to offer.}\\
        3.A: {\blue Do you have any dietary preferences or restrictions I should be aware of when recommending restaurants?}\\
        3.U: {\red I avoid eating spicy foods entirely.}\vspace{-0.2cm}
        \end{tcolorbox}
    \end{minipage}
    \\
    \begin{minipage}[t]{0.99\textwidth}
        \begin{tcolorbox}[colback=gray!5, colframe=gray!50, boxrule=0pt, arc=2mm]
        \vspace{-0.2cm}
        \textbf{Mistral-7B-Instruct}\\
        1.A: {\blue What type of cuisine are you most interested in trying during your visit to Mumbai?}\\
        1.U: {\red I'm interested in trying a variety of cuisines, but I'm particularly curious about authentic Indian dishes.}\\
        2.A: {\blue What specific Indian dishes would you like to try during your visit to Mumbai?}\\
        2.U: {\red I'm interested in trying dishes like butter chicken, biryani, and paneer tikka.}\\
        3.A: {\blue Based on your interest in authentic Indian dishes, have you tried any specific regional variations of Indian cuisine before? For example, there are notable differences between the cuisines of different Indian states, such as the Punjabi, Bengali, or South Indian cuisines. Knowing this could help me recommend some top-rated restaurants that specialize in those regional dishes.}\\
        3.U: {\red I've had the chance to try Punjabi cuisine before, and I've really enjoyed it. However, I haven't explored much of Bengali or South Indian cuisines yet, so I'm open to recommendations for those as well.}\\
        4.A: {\blue Based on your preference for Punjabi and openness to exploring Bengali and South Indian cuisines, can you specify if you have any dietary restrictions or preferences, such as vegetarian or vegan, gluten-free, or spicy food? This information could help me recommend restaurants that cater to your needs and preferences.}\\
        4.U: {\red I avoid eating spicy foods entirely. Other than that, I don't have any specific dietary restrictions.}\\
        5.A: {\blue Based on your preference for Punjabi cuisine and your avoidance of spicy foods, are there any specific Punjabi dishes that you would like to try during your visit to Mumbai that do not contain spicy ingredients? For example, certain Punjabi dishes can be prepared with mild or no spices. Knowing this could help me recommend top-rated restaurants that offer such dishes.}\\
        5.U: {\red I'm interested in trying dishes like butter chicken and paneer tikka that are prepared without spicy ingredients. If there are any other mild Punjabi dishes that you recommend, I'd love to know about those as well.}\vspace{-0.2cm}
        \end{tcolorbox}
    \end{minipage}
}
\caption{An instance of the Personalized Question Answering task for the ``Travel Restaurant topic'', along with the conversations of two models.}
\label{fig:qa_example2}
\end{figure*}

\section{Implementation Details}

\subsection{LLM Models}
In the experiments we used the following LLM models:
\begin{enumerate}
    \item GPT-4o-mini \cite{hurst2024gpt}
    \begin{itemize}
        \item Link: \url{https://openai.com/api/}
        \item License: proprietary (accessed through OpenAI API)
    \end{itemize}
    \item Claude-3.5-Haiku
    \begin{itemize}
        \item Link: \url{https://www.claude.com/platform/api}
        \item License: proprietary (accessed through Claude API)
    \end{itemize}
    \item Mistral-7B-Instruct \cite{jiang2023clip}
        \begin{itemize}
        \item Link: \url{https://huggingface.co/mistralai/Mistral-7B-Instruct-v0.2}
        \item License: Apache license 2.0
    \end{itemize}
    \item Qwen2.5-7B-Instruct \cite{yang2025qwen3}
    \begin{itemize}
        \item Link: \url{https://huggingface.co/Qwen/Qwen2.5-7B-Instruct}
        \item License: Apache license 2.0
    \end{itemize}
\end{enumerate}
The LLM models that were not accessed via APIs were run on NVIDIA A40 GPUs with 40GB of memory.

\section{Datasets}

\subsection{Data Sources and Generation}\label{sec:data_source}
Details about the data sources and the generation process are provided below. The data samples were manually inspected to ensure that they do not contain personally identifying info or offensive content.

\subsubsection{20 Questions Game}

The object data were generated using LLMs. Specifically, we provided GPT-4.1 with a set of five topics and requested it to generate twenty distinct objects per topic. 

\subsubsection{Personalized Question Answering}

For the Personalized Question Answering we used the data generation process and topics provided in \cite{zhaollms}. We selected five topics and used their code to generate ten question-preference pairs per topic. The code we used was released under a Attribution-NonCommercial 4.0 International license.

\subsubsection{Personalized Text Summarization}

In personalized text summarization we used text from four different text datasets. These are the following:
\begin{itemize}
\item News articles: \textit{ccdv/cnn\_dailymail}
\begin{itemize}
    \item Link: \url{https://huggingface.co/datasets/ccdv/cnn_dailymail}
    \item License: Apache license 2.0
\end{itemize}
\item General knowledge articles: \textit{wikimedia/wikipedia} 
\begin{itemize}
    \item Link: \url{https://huggingface.co/datasets/wikimedia/wikipedia}
    \item License: Creative Commons Attribution Share Alike 3.0
\end{itemize}
\item Science papers: \textit{heegyu/elsevier-oa-cc-by} 
\begin{itemize}
    \item Link: \url{https://huggingface.co/datasets/heegyu/elsevier-oa-cc-by}
    \item License: unspecified
\end{itemize}
\item Patent documents: \textit{satpalsr/bigpatent-test}
\begin{itemize}
    \item Link: \url{https://huggingface.co/datasets/satpalsr/bigpatent-test}
    \item License: unspecified
\end{itemize}
\end{itemize}

\end{document}